\documentclass[letterpaper, 10 pt, conference]{ieeeconf}  

\IEEEoverridecommandlockouts                              
 
\usepackage[style=ieee]{biblatex}
\usepackage[hidelinks,breaklinks=true]{hyperref}

\usepackage{graphicx} 
\usepackage{multirow}
\usepackage{booktabs}
\usepackage{tabularx}
\usepackage{float}
\usepackage{amsmath}

\title{\LARGE \bf
Supervised Autonomous Bimanual Suction and Cauterization for Bleeding Control in Low Grade Splenic Trauma
}

\author{  
    Francis Fan$^{1}$, 
    Mackenzie A Eagleson$^{2}$, 
    Yvonne Zhang$^{1}$, 
    Hongting Pan$^{1}$, 
    Lujia Yang$^{1}$,
    Ethan Kilmer$^{1}$, 
    Jiawei Ge$^{1}$, \\
    Nural Yilmaz$^{1}$,
    Avinash Dhimal$^{3}$, 
    Ahmed Ghazi$^{3}$, 
    Jeffrey Jopling$^{2}$,
    Laura Connolly$^{1 \dagger}$,
    and Axel Krieger$^{1 \dagger, *}$%
\thanks{$\dagger$~These authors contributed equally to this work.}%
\thanks{* Corresponding author: Axel Krieger (axel@jhu.edu)}
\thanks{$^{1}$Francis Fan, Yvonne Zhang, Hongting Pan, Lujia Yang, Ethan Kilmer, Jiawei Ge, Nural Yilmaz, Laura Connolly and Axel Krieger are with the Laboratory of Computational Sensing and Robotics (LCSR), Johns Hopkins University, Baltimore, MD 21218, USA}
\thanks{$^{2}$Mackenzie A Eagleson and Jeffrey Jopling are with the Department of Surgery, Johns Hopkins Medicine, Baltimore, MD 21287, USA}
\thanks{$^{3}$Avinash Dhimal and Ahmed Ghazi are with the Brady Urological Institute, Johns Hopkins University, Baltimore, MD 21287, USA
}
}

\begin{document}

\maketitle
\thispagestyle{empty}
\pagestyle{empty}

\begin{abstract}

Uncontrolled hemorrhage remains a leading cause of preventable trauma mortality. Timely intervention is not always available, making autonomous systems that can aid in bleeding control a promising alternative. Given the safety-critical nature of the task, supervised autonomy,  in which the system executes the technical steps while a human retains oversight,  represents a necessary intermediate step toward this goal. In this work, we present a supervised autonomous bimanual robotic framework for injury cauterization (eletrosurgical coagulation) in low-grade splenic trauma. The system integrates real-time bleeding segmentation, markerless tool keypoint tracking, and deep tool-tissue contact detection to coordinate two robotic arms for synchronized suction and cauterization. Evaluated on a novel, cannulated ex vivo splenic hemorrhage model, the system achieved an average bleeding surface cauterization coverage of 82.9\% across seven distinct injuries. This work demonstrates the feasibility of using a supervised autonomous bimanual system to deliver targeted coagulation across an injury site.

\end{abstract}

\section{INTRODUCTION}

Uncontrolled bleeding, otherwise referred to as hemorrhage, is the leading cause of preventable death in modern armed conflict. An analysis of US battlefield deaths between 2001 and 2011 discovered that 87.3\% of injury mortality occurred before reaching a medical treatment facility (MTF). Among potentially survivable pre-MTF deaths, 90.9\% were associated with hemorrhage \cite{eastridge2012death}. The most lethal presentation is non-compressible truncal hemorrhage, defined as severe torso bleeding that cannot be controlled by external compression and for which no adequate field intervention is currently available. The same problem arises in civilian surgical settings. While intraoperative death is now rare, when it occurs, uncontrolled bleeding is the leading proximate cause \cite{gallastegi2022intraoperative}. In a prospective cohort study, even in a controlled OR environment, bleeding was estimated to account for 20–32\% of all 30-day postoperative deaths \cite{roshanov2021bleeding}.

Following traumatic injury, survival depends on achieving hemostasis and delivering definitive intervention as quickly as possible \cite{abhilash2020early}. However, definitive care is bottlenecked by whether trained personnel are available and by how quickly they can localize a bleed and control it. Autonomous medical robots offer a path to decoupling this care from both human proximity and human speed, ultimately shortening time to hemostasis in settings where a provider is available, and extending intervention to settings a provider likely cannot reach in time.

\begin{figure}[t]
    \centering
    \includegraphics[width=0.95\linewidth]{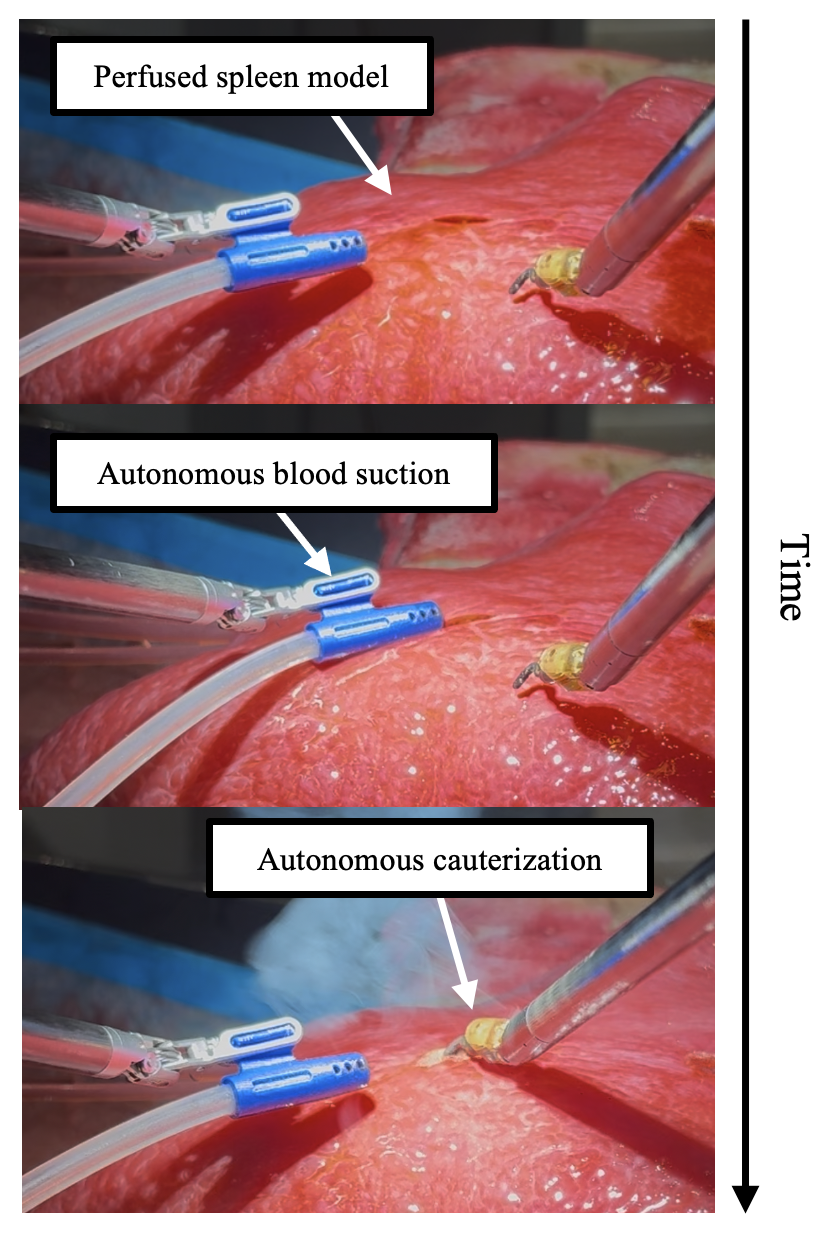}
    \caption{Supervised autonomous bimanual suction and cauterization for bleeding control. Images progress in time from top to bottom. \emph{Top:} Blood pooling from a simulated splenic injury. \emph{Middle:} Autonomous suction clearing blood from the injury site. \emph{Bottom:} Autonomous cauterization covering the injury surface to initiate bleeding control.}
    \label{fig:placeholder}
\end{figure}

\begin{figure*}[t]
    \centering
    \includegraphics[width=1.0\linewidth]{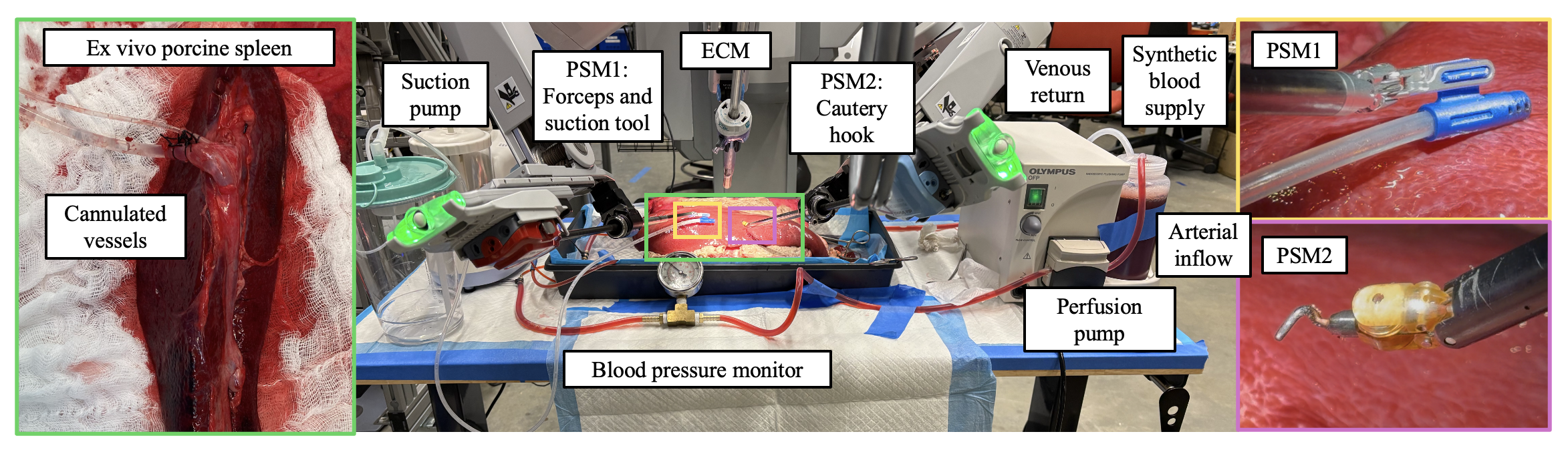}
    \caption{System diagram: \emph{Left:} Cannulated vessels in the \textit{ex vivo} splenic hemorrhage model. \emph{Middle:} Experimental setup including suction pump, robotic tools, and perfusion system. The splenic hemorrhage model is outlined in green; robotic instrument positions are highlighted in yellow and pink. \emph{Right, top:} Close-up of the yellow-outlined tool, a custom-designed robotic suction tool fitted to ProGrasp forceps. \emph{Right, bottom:} Close-up of the pink-outlined tool, a cautery hook instrument used for wound coagulation.}
    \label{fig:system_diagram}
\end{figure*}

Realizing this vision of autonomous intervention first requires a robot that can identify where blood is pooling and where it is coming from. There have been efforts to augment human perception for hemorrhage control through real-time bleeding segmentation and bleeding source identification. For example, Okamoto et al. used a support vector machine to identify bleeding regions and provide an overlay to the surgeon to assist with rapid packing of the wound \cite{okamoto2019real}. More recently, Xu et al. proposed BleedOrigin-Net, an architecture that can be used to identify a bleeding source and track its position over time for endoscopic submucosal dissection \cite{xu2025bleedorigin}. Similarly, Pei et al. proposed BlooDet, a dual-task model that can produce a segmentation mask of bleeding regions and their source location \cite{pei2026synergistic}. Sogabe et al. proposed the Bleeding Alert Map, a model trained on data from an injury-mimicking device to map bleeding images to their corresponding bleeding sources \cite{sogabe2023bleeding}. There have also been many efforts to produce clinical datasets that will support training of robust perception models for bleeding detection such as Hemoset \cite{miao2024hemoset}, Surgblood \cite{pei2026synergistic}, and BleedOrigin-Bench \cite{xu2025bleedorigin}. Each of these approaches has demonstrated that it is possible to train a model to identify bleeding pools and sources, however, none have coupled these perception techniques with autonomous robot action. Closing this perception-to-action gap requires first understanding what physical intervention autonomous hemostasis actually demands.


Depending on the degree of injury, hemostasis control relies on four key maneuvers: suction, cauterization (electrosurgical coagulation), packing, and ligation. Suction and packing are used to soak up and remove pooling blood so the surgeon can identify the bleeding source, while cauterization and ligation are used to close the wound and prevent persistent bleeding. Achieving autonomous hemostasis therefore requires a robot capable of reasoning about and executing across all four of these maneuvers, not just one in isolation.

Of these four maneuvers, autonomy research to date has focused almost exclusively on optimizing the trajectory of robotic suction tools based on pooling blood. Richter et al., for example, developed a probabilistic blood flow detection system that guides the trajectory of a robotic suction tool \cite{richter2021autonomous}. They detect flowing blood using optical flow with temporal filtering and devise a suction trajectory technique that determines beginning and end points for suction based on estimates of the oldest and newest blood in the scene. While promising, the technique assumes that the only motion in the surgical environment is caused by fluids or surgical tool movements, when in practice, patient respiration can cause additional movement that would make optical flow techniques challenging. Hua et al. then combined optical flow with a region-based CNN model to capture spatial features and compensate for these limitations \cite{hua2022automatic}. Simulation-based blood flow modeling has also been demonstrated successfully. Huang and Liu et al. incorporated differential fluid simulations for suction into a model-predictive controller to determine suction tool trajectories \cite{huang2021model}. Ou et al. instead used position-based fluid simulation to train a generalizable robotic suction policy via reinforcement learning \cite{ou2024autonomous}. More recently, this work was extended to include a large language model (LLM) for task reasoning to guide the RL-based autonomous robotic suction \cite{zargarzadeh2025decision}. A simulation platform called CreSSim-ML, built for simulating the da Vinci Research Kit (dVRK), has similarly been used for autonomous irrigation and suction \cite{ou2025learning}. Together, these works address only one of the four maneuvers required for hemostasis, leaving cauterization, packing, and ligation largely unautomated. Moreover, most are validated on synthetic phantoms or simplified bench-top setups rather than in clinically representative conditions, revealing a significant realism gap in how these systems are tested.

In this paper, as a step toward autonomous bleeding control, we present a supervised, bimanual robotic system for bleeding control that incorporates both suction and cauterization. Our system is designed to detect an injury, use robotic suction to clear the blood, and then cauterize the injury to close it. We use a segmentation model to identify pooling blood, a markerless tool tracking model for visual servoing, and a perception model to identify tool-tissue contact. We validate the system on a novel, perfused, \textit{ex vivo} splenic hemorrhage model because the spleen is a common and clinically significant source of hemorrhage in blunt abdominal trauma, and the organ's high vascularity closely mimics human bleeding conditions. The contributions of this work include: 
\begin{enumerate}
    \item A perfused, \textit{ex vivo} splenic hemorrhage model for realistic data collection, validation, and testing
    \item A vision-only perception pipeline that localizes injuries, tracks instruments, and infers contact from visual cues in laparoscopic video
    \item Supervised autonomous robotic bimanual suction and cauterization for bleeding control 
\end{enumerate}

\section{METHODS}

Figure \ref{fig:system_diagram} presents the experimental setup and robotic system. The da Vinci Research Kit (dVRK) Si system was used for all data collection and autonomous experiments \cite{xu2025dvrk}. For this demonstration we use two patient side manipulators (PSM) and the endoscopic camera manipulator (ECM). PSM1 was equipped with a Prograsp Forcep tool, fit with a custom 3-D printed suction device. PSM2 was equipped with a permanent cautery hook.  

\subsection{Ex vivo splenic hemorrhage model} 
We developed an \textit{ex vivo} hemorrhage model by cannulating the splenic artery and vein of porcine spleens, allowing for reproducible perfusion, pressurization, and graded injury testing under controlled physiologic-analog conditions. Data collected from this hemorrhage model was used for autonomy training and validation. 

\subsubsection{Back-table porcine spleen cannulation} Spleens were purchased from Animal Technologies or procured during surgical resident education following swine euthanasia. Surrounding perisplenic tissue was removed and specimens were inspected for injury. Branches off the splenic vessels were identified and suture-ligated with 2-0 silk to prevent leak. 

The splenic artery and vein were then dissected under magnification. Approximately 3~cm of both vessels were mobilized to permit cannula placement. Appropriately sized arterial (2.0-3.5~mm OD/1.0-2.5~mm ID) and venous (3.5-5.0~mm OD/2.5-3.5~mm ID) cannulas were inserted and advanced toward the splenic hilum without passing the first vascular bifurcation. Each cannula was secured with three 2-0 silk ties to achieve a watertight seal. The prepared spleen was then kept on ice until experimentation.

\subsubsection{Synthetic blood} A blood-mimicking fluid was developed as the perfusate in our \textit{ex vivo} hemorrhage model. The formula was constructed to: (1) conduct current adequately for electrocautery, (2) contain proteins that irreversibly denature and visibly coagulate, and (3) mimic blood's optical appearance. The synthetic blood is comprised of: pasteurized liquid egg whites at 5g/dL (normal human plasma concentration) as our thermally coagulable protein source, 0.9\% sodium chloride for ionic conductivity, glycerol 5\% by volume to approximate blood viscosity (3cP), and Red No. 40 and Blue No.1 food dye for visual appearance. The perfusate was heated to $37\,^{\circ}\mathrm{C}$ prior to experimentation.

\subsubsection{Perfusion circuit \& experimental trauma} A closed-loop circuit was established from the reservoir containing our synthetic blood. Synthetic blood was infused into the arterial cannula via a manual roller pump (Olympus OFP-2) with an in-line pressure transducer, and returned out the venous cannula via passive drainage back to the reservoir. Circuit components were primed and de-aired prior to perfusion. At low flow, the spleen was carefully inspected for vascular injuries; any leaks identified were repaired with 6-0 Prolene. Flow was gradually increased and titrated to a target perfusion pressure of 65–110 mmHg (1.25–2.13 PSI). The circuit was confirmed to maintain target pressure without uncontrolled leakage prior to initiating experimental trauma.

Splenic lacerations were created using a \#11 blade. In accordance with the American Association for the Surgery of Trauma (AAST), the depth of laceration defines the grade of splenic injury, with wounds $<$ 1~cm in depth defined as grade I and wounds 1-3~cm in depth defined as grade II. To facilitate uniform, reproducible injuries across AAST grades, the surgical blade was marked at 1~cm intervals as depth reference. This technique successfully produced graded parenchymal defects and bleeding rates. Monopolar energy was applied to the cautery hook using a DRE ASG-300 generator. Energy mode was set to fulguration or ``spray” (high voltage, interrupted duty cycle) and power titrated to tissue effect from 35-60W.

\subsection{Overall bimanual control workflow}

\begin{figure}[!h]
    \centering
    \includegraphics[width=1.0\linewidth]{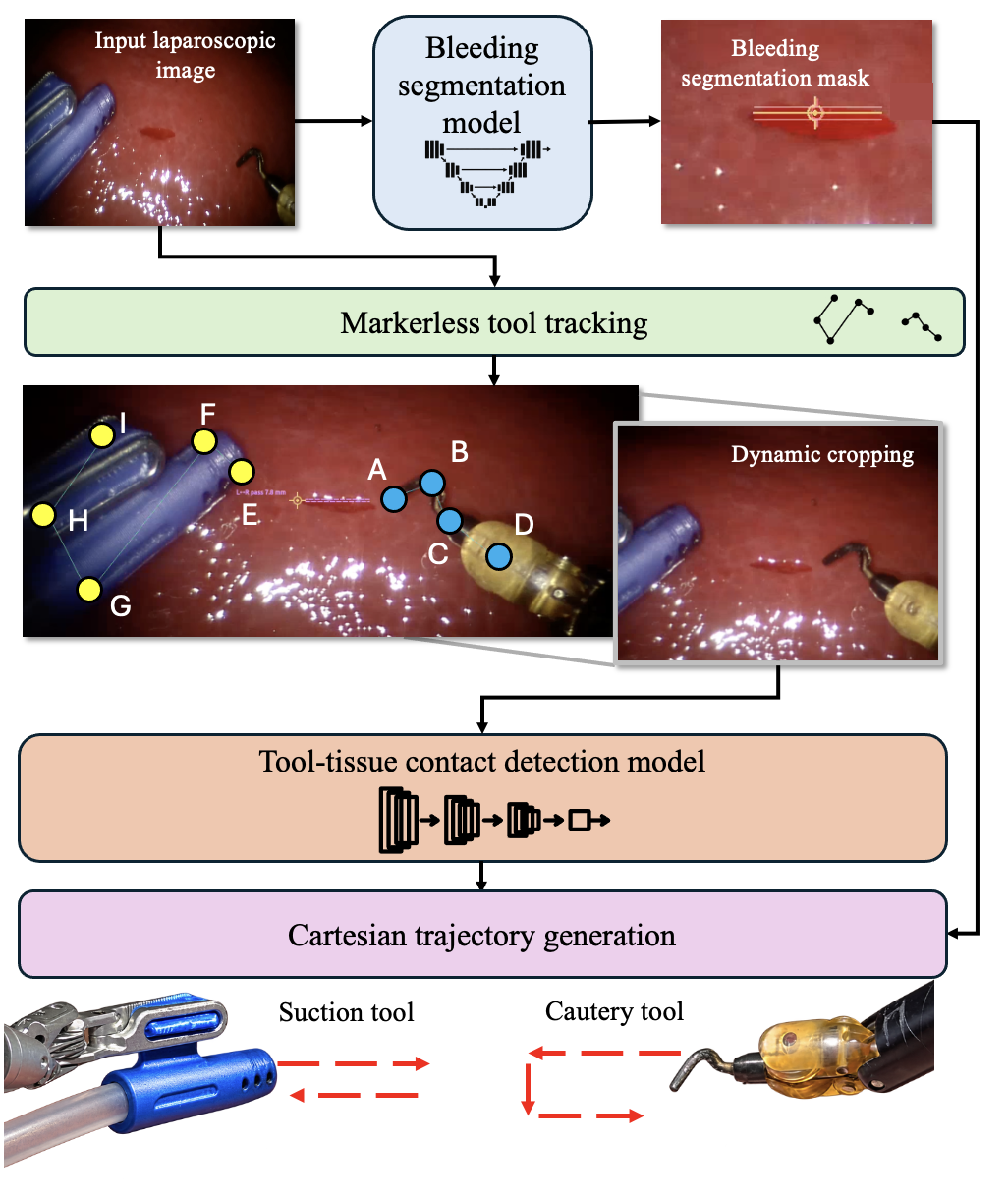}
    \caption{Overview of the supervised autonomous bimanual suction and cauterization workflow. The bleeding segmentation model is run on a laparoscopic image to produce a bleeding segmentation mask, from which the injury location is derived. The positions of both robotic tools (suction and cautery) are extracted from the image using a markerless tool-tracking framework based on keypoint detection. This tracking information is used to crop the image to a smaller region of interest for input to the tool–tissue contact detection model. The robotic tools are moved to the injury location via visual servoing, lowered until contact is detected, and then used to execute an autonomous Cartesian trajectory along the wound length.}
    \label{fig:overall_model}
\end{figure}

Our approach uses perception models to localize the injury and robotic instruments, then determine tool-tissue contact, enabling coordinated bimanual suction and cauterization. The trajectory of each tool is derived from the length of the injury and location in the image. Figure \ref{fig:overall_model} provides an overview of the perception models and robot control pipeline. The following sections outline each component. 

\subsection{Perception models}
The perception pipeline is broken down into three  models: (1) a bleeding segmentation model to localize the injury, (2) a markerless tool tracking model for visual servoing, (3) a tool-tissue contact detection model to determine when the tools are in the correct locations for blood suction and injury cauterization. 

\subsubsection{Bleeding segmentation model} While the focus of this paper is on bimanual robotic suction and cautery, we trained a simple segmentation model to detect pooled blood to help localize injuries. An nnU-Net was selected for this task due to its self-configuring nature: given only the raw dataset and target labels, it automatically determines key design choices, including image preprocessing (e.g., resampling and normalization), network architecture and topology, patch size, and training hyperparameters, without requiring manual tuning \cite{isensee2021nnu}. This data-driven approach to configuration has been shown to perform well on medical image segmentation tasks \cite{isensee2024nnu}. For this application, the framework configured a 7-stage 2D U-Net with a batch size of 20 and a patch size of 320×512. The network was trained on the training set for 250 epochs using SGD with Nesterov momentum (0.99), an initial learning rate of 0.01 with polynomial decay, and a combined Dice and cross-entropy loss, on an NVIDIA GeForce RTX 5080 GPU with 16 GB of VRAM.

The network output is then reduced to a geometric estimate of the injury under the assumption that injuries appear horizontally across the surgical view with an approximately ellipsoidal shape. Predicted masks are smoothed by majority vote over the last three frames, and the rows of the highest-reaching component are scanned downward from its apex. The wound's upper edge is taken as the first row whose blood-pixel count reaches 95\% of the component's characteristic width and holds for four consecutive rows, marking the transition from the rounded cap to the wider body. This row is spanned from its leftmost to its rightmost blood pixel, and a second span of equal width is mirrored the same distance below it to approximate the wound's lower extent. The median x position of the upper edge is published as the candidate bleeding source. 







\subsubsection{Markerless tool tracking} %
DeepLabCut (DLC) was used for markerless 2-D tracking of the surgical instruments \cite{Mathisetal2018} \cite{NathMathisetal2019}. The tracker provides instrument geometry for visualization, control, as well as constructing dynamic image cropping region to be used by the contact detector. In a single frame, nine distinctive tool keypoints across two tools are detected: four distinct keypoints on the cautery hook (Figure \ref{fig:overall_model}: A-D), and five distinct keypoints on the suction tool (Figure \ref{fig:overall_model}: E-I).



While most of the points are visible to the endoscope during the surgical tasks, change in illumination conditions, and artifacts such as blood and smoke might cause the confidence of the keypoints to drop. To mitigate this, fall-back points are selected to substitute the primary control points: when the primary keypoint \emph{A} has confidence lower than 0.6 the \emph{B} is chosen for the suction tool; when the confidence for \emph{E} falls under 0.6, \emph{F} is chosen for the monopolar hook. Due to their close proximity to their corresponding primary control points, the functions of the tools are unaffected even if the fall back happens.


The learned keypoints also provide a geometric representation of the instrument pose. Because the keypoints are rigidly connected by the tool geometry, distances between them can be used as stable image-space measurements. When the instrument and tissue surface are approximately coplanar, a local pixel-to-metric scale can be estimated from a reference length on the tissue surface:
\[
s_{\mathrm{tissue}} =
\frac{L_{\mathrm{reference}}^{\mathrm{mm}}}
     {d_{\mathrm{reference}}^{\mathrm{px}}}
\quad [\mathrm{mm/px}],
\]
where $L_{\mathrm{reference}}^{\mathrm{mm}}$ is a known physical reference length and $d_{\mathrm{reference}}^{\mathrm{px}}$ is its image-space length. A keypoint distance can then be converted approximately to metric units as
\[
L_{\mathrm{keypoints}}^{\mathrm{mm}} =
d_{\mathrm{keypoints}}^{\mathrm{px}}
s_{\mathrm{tissue}}.
\]
The same scale can be applied to image-space trajectories or wound-length measurements when the relevant points remain near the tissue plane. These measurements are approximate because monocular image scale varies with depth, perspective, camera motion, and tissue deformation; therefore, this estimation is based on the assumptions that the tool and tissue surface is on approximately the same depth plane.

The keypoint detector was trained using DLC 3.0.1 with the PyTorch backend and a ResNet-50 pose-estimation backbone. 
The model was trained for 200 epochs using a staged learning-rate schedule on an NVIDIA GeForce RTX 5070 GPU with 12~GB of VRAM. 


During inference, DLC returns an $(x,y)$ coordinate and a confidence value for every keypoint. Primary control points are accepted only when their confidence exceeds the configured gate; otherwise, the corresponding fallback point is used when available. Reliable points from both tools are then combined to construct a dynamic frame crop for the contact classifier. The crop is expanded by a fixed margin and converted to the classifier's target aspect ratio. Its center and size are temporally smoothed, and the previous valid crop is briefly retained when a frame contains low-confidence or missing keypoints. This prevents isolated DLC failures from producing abrupt crop motion or a transient loss of the contact-classifier input.

\subsubsection{Tool-tissue contact detection model} 

We trained a binary tool--tissue contact detection model using an EfficientNet-B0 backbone \cite{tan2019efficientnet} to detect physical contact on the tissue surface. The original classification head was replaced with a two-class linear layer (contact vs. no-contact). To match deployment conditions, candidate training frames were processed through the DLC pipeline to extract the dynamic dual-tool crops described above. Frames with low keypoint confidence or simultaneous two-tool contact were excluded. The resulting crops were grouped by recording session for grouped 5-fold cross-validation, normalized using ImageNet statistics, resized to $224 \times 168$ pixels, and padded to $224 \times 224$. Training augmentation included geometric perturbations and photometric variation, including translation, scaling, rotation, color jitter, brightness variation, and contrast variation.

The model was evaluated using grouped five-fold cross-validation. Training used batches of 128 samples, automatic mixed precision, and an NVIDIA GeForce RTX 5070 GPU with 12~GB of VRAM. The learning rate was initialized at $10^{-4}$ and decayed during training. Validation performance was monitored using the contact F1 score, defined as $F_1=2PR/(P+R)$, where precision (\emph{P}) is the fraction of predicted contact frames that were correct and recall (\emph{R}) is the fraction of true contact frames that were detected. Early stopping and validation-based checkpoint selection were used to reduce overfitting: training ended after five consecutive epochs without a higher validation contact F1, and the checkpoint with the highest validation F1 was retained for each fold. The final deployable checkpoint was then trained on all accepted crops using the median number of selected epochs across the five folds.

\subsubsection{Training datasets}
To train each of the perception models, we generated data from simulated injuries on 2 \textit{ex vivo} spleens and recorded an expert surgeon treating each injury using the same instruments as the supervised autonomous robotic system, with video captured from the da Vinci Si endoscope camera. We refer to each recording as an ``injury treatment recording". Using the Computer Vision Annotation Tool (CVAT) \cite{sekachev2020cvat}, we generated three datasets from this footage: (1) a bleeding segmentation dataset, (2) a markerless tool tracking dataset, and (3) a contact detection dataset.

\emph{\textbf{Bleeding segmentation dataset:}} Pooling and pooled blood in close proximity to the surgical tools were manually annotated according to the guidelines described by Miao et al. \cite{miao2024hemoset}. In total, 7 injury treatment recordings were used to construct the bleeding segmentation dataset. Videos were sub-sampled at a rate of 1 frame per 10 seconds and split into training, validation, and test sets. In total, 3525 frames are used for training, 210 are used for validation, and 330 are used for testing. 


\emph{\textbf{Markerless tool tracking dataset:}} The markerless tool tracking dataset was extracted from 23 injury treatment recordings. Frames were selected from each recording to capture representative instrument poses and challenging visual conditions, including partial occlusion, blood, smoke, and changes in illumination. The selected frames were manually annotated with keypoints. In total, 2,529 frames were annotated, with up to nine keypoints per frame distributed across the suction tool and the cautery hook. The annotated data were then split such that 95\% (2,402 frames) were used for training and the remaining 5\% (127 frames) were used for testing. Since the markerless tool tracking model focuses on the tools themselves rather than scene-level context, frames were split randomly rather than by recording.

\emph{\textbf{Contact detection dataset:}} The contact detection dataset was constructed from 21 injury treatment recordings, supplemented with additional recordings captured by non-experts in which tool–tissue interactions were deliberately performed under challenging visual conditions, including varying scene illumination, injury geometry, bleeding intensity, instrument poses, and appearance changes throughout the procedure. Each frame was assigned a binary label: contact if either instrument touched the tissue, and no contact otherwise. Frames with simultaneous contact by both instruments were excluded, since our control system operates only one instrument at a time.

In total, $93\,893$ frames were labeled: $60\,489$ as no-contact and $29\,887$ as single-tool contact. The dataset was partitioned by complete recording rather than by individual frame to avoid leakage from shared scene geometry. We used grouped five-fold cross-validation, with each fold holding out complete recordings so that temporally adjacent frames from the same recording could not appear in both training and validation sets.


\subsubsection{Online perception deployment}

The supervised autonomous system receives a $1920\times1080$ video stream at $30\text{~Hz}$ from the left endoscopic camera of the da Vinci Si. The same stream is provided to the three perception models, each with its own processing resolution and update rate; these resizing operations do not alter the camera stream itself. For bleeding detection, each incoming frame is resized to $480\times270$ and processed by nnU-Net at approximately $3\text{~Hz}$. The network produces a binary bleeding mask, from which three guide lines are computed to describe the wound geometry. For instrument tracking, frames are resized to $1280\times720$ before being processed by DLC at approximately $7\text{~Hz}$. DLC estimates the image coordinates and confidence values of nine tool keypoints across the suction tool and cautery hook.

The tool-tissue contact detection model operates on every available video frame at approximately $30\text{~Hz}$. The most recent valid DLC keypoints are used to define a single dynamic crop spanning the two instruments. The crop is expanded by a fixed margin, converted to the contact classifier's target aspect ratio, and resized to the classifier input size of $224\times224$ pixels before classification. Thus, DLC determines the crop location at its lower frame rate, whereas the contact detection model repeatedly classifies the newest video frames using the most recently available crop at a faster frame rate. The two update loops are decoupled so that a temporary delay in keypoint tracking does not queue or delay video frames. The crop location is temporally smoothed and the previous valid crop is retained briefly when keypoint confidence is low, reducing sudden crop motion while preserving continuous contact predictions.

\subsection{Robot control}

The online perception outputs are used to inform autonomous robot actions. The following sections describe the action breakdown in more detail.


\subsubsection{Image-Jacobian Visual Servoing}

To control lateral tool motion, pixel residuals $e = [u_{\text{target}} - u_{\text{proxy}}, v_{\text{target}} - v_{\text{proxy}}]^T$ are mapped to physical displacements $\Delta x, \Delta y$ using local image Jacobians $\mathbf{J}$ calibrated for both primary and fallback keypoints across the PSM1 and PSM2 instruments. The lateral control update vector $\delta$ (in mm) is computed using a damped least-squares inverse formulation, given by:

\begin{equation}
\delta = \text{gain} \cdot \mathbf{J}^T \left( \mathbf{J} \mathbf{J}^T + \lambda^2 \mathbf{I} \right)^{-1} e
\end{equation}

\noindent where the controller schedules lower gain and damping ($\lambda$) parameters near the target, applies exponential smoothing ($\alpha \approx 0.55$) to the target residual, and enforces a $30\text{ mm}$ maximum total lateral travel limit via bounded $25\text{ Hz}$ cartesian setpoint commands (\texttt{servo\_cp}). Convergence is confirmed when the image residual remains within an $8\text{ px}$ threshold for $1.5\text{ s}$ under valid Jacobian and tracking conditions, with the action pausing for up to $3\text{ s}$ or initiating a $3\text{ mm}$ visibility probe before failing closed if DLC tracking is temporarily lost.

\subsubsection{Depth Control and Injury Approach}

Once lateral alignment is verified, depth control operates in a contact-only mode without requiring absolute 3D depth estimations, commanding bounded axial motion at $0.75\text{ mm/s}$ ($25\text{ Hz}$) up to a $40\text{ mm}$ maximum travel limit. To compensate for tissue deformation or relative motion during descent, the controller dynamically recenters lateral position whenever the target residual exceeds $10\text{ px}$, capping corrections at $0.20\text{ mm}$ per update, and halts motion upon suspect ($>$0.6 contact probability) or confirmed ($>$ 0.8 contact probability) tool--tissue contact, stale inference, or a $3.0\text{ s}$ timeout.


\subsubsection{Supervised Autonomous Workflow}


The high-level execution framework governs three operational phases: preflight checks, autonomous planning, and coordinated dual-tool autonomy. As a safety precondition, autonomous commands remain disabled until recovery points have been saved for both patient-side manipulators. Before entering the autonomous phase, a supervisor confirms that the estimated injury segmentation is accurate and activates autonomous suction and cautery; the system then freezes the segmentation and trajectory geometry for the remainder of the procedure. PSM1 descends to an entry point on the wound centerline, located $1\text{ mm}$ to the left of the centerline's leftmost point, locks its scale, and performs a centerline suction sweep, proceeding toward the right side of the wound for $75\%$ of the estimated wound length before returning left at the same speed. While PSM1 performs the rightward sweep, PSM2 positions the monopolar hook and applies image-based $x/y$ corrections to align with the treatment target. Because smoke can obscure the visual features required for reliable servoing, PSM2 then switches to precomputed, fixed-orientation \texttt{servo\_cp} treatment strokes at a default speed of $1\text{ mm/s}$, lifting to a $2\text{ mm}$ clearance height between passes before descending again for the next stroke. Energy activation remains manually gated: the supervisor instructs an operator at the operating table to press the monopolar pedal during each treatment pass. Autonomous energy activation is reserved for future system iterations.


\section{EXPERIMENTS \& RESULTS}

\subsection{Bleeding segmentation model testing}

We evaluated the performance of the nnU-Net on the annotated injury treatment recordings reserved for testing from the bleeding segmentation dataset. The model detected bleeding in 98\% of frames in which it was present (n = 297), achieving a median Dice of 0.69 and median HD95 of 36.6 px. Sensitivity (0.77) exceeded precision (0.58), indicating a tendency toward over-segmentation relative to the manual annotations. While the model reliably localizes the injury region, predicted segmentation masks occasionally underestimate or overestimate the extent of the injury or pooled blood.





\subsection{Markerless tool tracking testing}

The markerless tool-tracking model was evaluated on 127 held-out frames. It achieved a mean average precision (mAP) of $93.76\%$ and a mean average recall (mAR) of $95.51\%$. The RMSE was $4.28$ pixels when considering predictions with confidence greater than $0.6$. Each frame contained up to nine annotated keypoints distributed across the cautery hook and suction tool.

The model performed reliably across most scenes for both instruments. The principal failure cases involved the cautery hook during coagulation, where smoke, electrical arcing, and partial immersion in pooled blood reduced keypoint confidence. Suction tool tracking was generally more stable, although confidence decreased at steep wrist angles relative to the camera. 

\subsection{Contact detection model testing}


The contact detection model was evaluated using grouped five-fold cross-validation, with complete recordings held out in each fold. Across the five folds, the model achieved a mean contact F1 score of $0.738$, with fold-wise values ranging from $0.691$ to $0.782$. 

The binary classifier performed reliably in dry scenes and during clear tool--tissue interactions. Performance was less consistent on wet, blood-covered tissue. Blood pools often produce a relatively textureless surface with weak visual changes when a tool approaches or contacts the tissue. In these cases, the model occasionally assigned a high contact probability while the tool was still hovering above the blood pool. This behavior suggests a scene-specific bias in the training data. In the injury-treatment recordings collected by an expert surgeon, the tools generally approached the blood pool laterally and were already close to the tissue before contact occurred, whereas the autonomous controller approaches the tissue more vertically. Because these recordings comprise a substantial portion of the training dataset, the classifier may associate the appearance of a nearby tool and blood pool with contact even when physical contact has not yet occurred.



\subsection{Bimanual suction and cauterization trials} 

Finally, to assess system performance on bimanual suction and cauterization, we conducted experiments on an unseen spleen with 5 simulated grade I injuries and 2 simulated grade II injuries. Autonomous suction and cautery activation waited for supervisory approval of the segmentation mask and injury localization before proceeding with the procedure. The injuries were evaluated after the cauterization path was completed to assess performance.



To quantify cauterization efficacy, percentage of bleeding surface that was cauterized is calculated from the pixel-level overlap between the target wound mask ($M_W$) and coagulation mask ($M_C$):
\begin{equation}
\text{Coverage (\%)} = \frac{|M_W \cap M_C|}{|M_W|} \times 100
\end{equation}
For qualitative evaluation (Fig.~4), region overlays are categorized as covered target ($M_W \cap M_C$, green), untreated wound ($M_W \setminus M_C$, red), and off-target coagulation ($M_C \setminus M_W$, yellow).

\begin{figure*}[!h]
    \centering
    \includegraphics[width=1.0\linewidth]{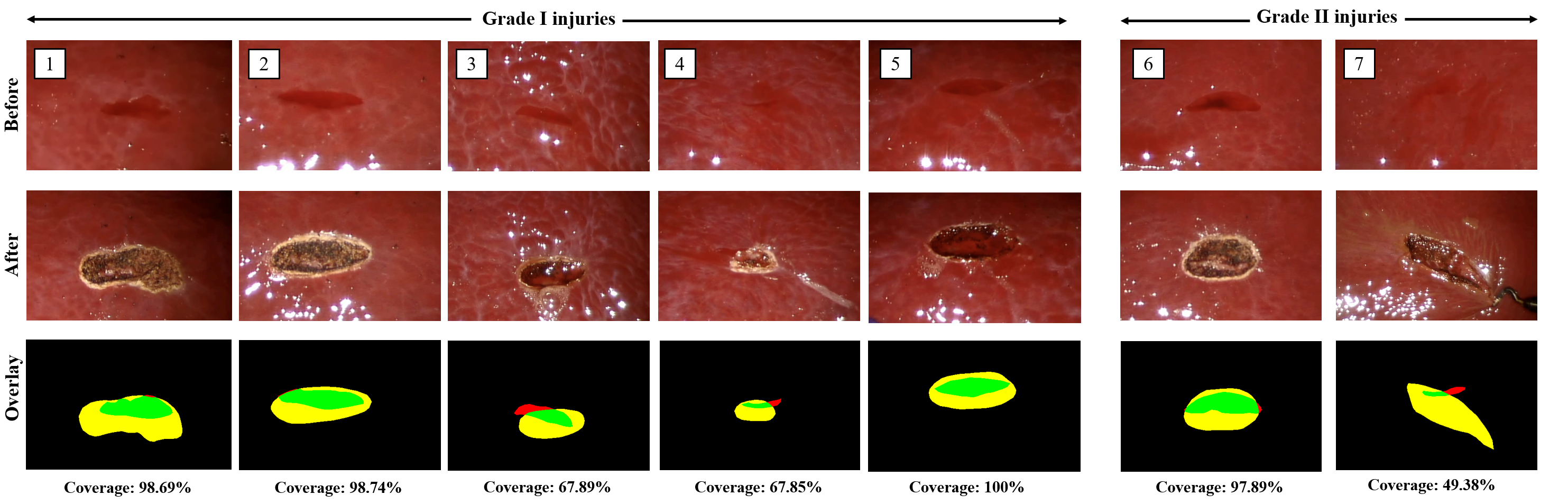}
    \caption{Images of injuries before and after suction and cauterization. \emph{Top row:} Injury before any intervention. \emph{Middle row:} Injury after cauterization. \emph{Bottom row:} Overlap between the injury mask and the cauterized surface area.  }
    \label{fig:coverage_figure}
\end{figure*}

\begin{table}[!t]
\centering
\caption{Results of supervised autonomous bimanual suction and cautery. Column 1 is the injury ID, corresponding to Figure \ref{fig:coverage_figure}. Columns 2 - 4 represent the extend of the injury (grade, length). Column 5 is the percentage of bleeding surface area cauterized.}
\label{tab:results_hemostatic_control}
\renewcommand{\arraystretch}{1.2}
\small
\newcolumntype{Y}{>{\centering\arraybackslash\hsize=0.85\hsize}X}
\newcolumntype{Z}{>{\centering\arraybackslash\hsize=1.6\hsize}X}
\begin{tabularx}{\columnwidth}{@{}YYYYZ@{}}
\toprule
\textbf{Injury} & \textbf{Grade} & \textbf{Depth} & \textbf{Length} & \textbf{Coverage} \\
 & & \textbf{(mm)} & \textbf{(mm)} & \textbf{(\%)}  \\
\midrule
1 & I  & 10 & 7.4 & 98.69 \\
2 & I  & 10 & 7.8 & 98.74 \\
3 & I  & 10 & 8.4 & 67.89 \\
4 & I  & 10 & 5.2 & 67.85 \\
5 & I  & 10 & 10.0 & 100 \\
6 & II & 20 & 7.6 & 97.89 \\
7 & II & 20 & 6.0 & 49.38 \\
\bottomrule
\end{tabularx}
\end{table}

Table \ref{tab:results_hemostatic_control} shows how these results correlate to injury length and severity. We achieve an average surface area coverage of 82.9\%. 4/7 injuries were cauterized with 97\% or more coverage. On Injury 3, 4, and 7 the coverage achieved was less than 70 \% which is likely due the surface geometry. The cautery hook must be held just above the injury to successfully cauterize. Because our perception pipeline lacks 3D surface reconstruction, we model the motion plane as orthogonal to the camera's optical ($z$) axis. Consequently, the planned trajectories cannot adapt to significant depth variations along the wound profile. We also observed two clear failure modes on injury 4 and injury 7 (Figure \ref{fig:coverage_figure}). On injury 4, tissue scarring occurred as the cautery hook exited the wound, both due to manual timing of the cautery pedal and the tool motion trajectory near the surface. For injury 7, where it has a deeper depth at grade II, the cautery hook got stuck inside the laceration. 


\section{DISCUSSION}

The bleeding segmentation model provided injury masks that supported supervised autonomous treatment, although, broader training data covering additional surgical scenes, tissue appearances, and injury geometries will be needed to improve generalization. The current geometric heuristic for estimating the injury extent from the segmentation mask is also a limitation. Since prior work has demonstrated more precise bleeding-source localization from laparoscopic video \cite{pei2026synergistic}, future versions will replace this heuristic with a learned estimate.

The main perception and control limitations involved markerless tool-tracking latency and contact-classification bias. DLC tracking required downsampling, which increased the keypoint update rate, but the resulting perception loop remained slower than the video stream. This slow inference limited the speed of autonomous actions and constrained our ability to assess the system under continuous blood flow conditions. In future work, we will optimize model inference to run the perception models at a higher frequency, allowing us to use the hemorrhage model to its full capacity.

Keypoint tracking confidence also occasionally decreased when the suction tip was submerged in blood or visually stained. The monopolar hook was generally easier to track because of its simpler visible geometry; however, smoke, arcing, and partial occlusion still reduced confidence during energized treatment. Additional annotated keypoints for challenging scenes could support future occlusion handling, approximate monocular depth estimation, and more complete wrist-pose recovery.

The contact classifier performed reliably in many dry and well-textured scenes but exhibited false-positive contact predictions above textureless blood pools. This behavior was most apparent during vertical tool descent stage, where the tool can occlude the tissue without physically contacting it. The failure is consistent with a distribution mismatch between the training data, which predominantly contains lateral human-operated approaches, while the autonomous motion includes more top-down approaches. This mismatch caused instability in the model when the tool is occluding the wound area and hovering. Additional training examples containing shallow contact, hovering above blood, reflections, and vertical descent should improve robustness.

The absence of explicit three-dimensional surface reconstruction also limited treatment of deeper Grade II injuries, where planar trajectory assumptions occasionally caused the cautery hook to snag or fail to follow the tissue surface. In the present experiments, treatment performance was summarized using surface cauterization coverage, which achieved 82.9\% on average for Grade I--II injuries. This metric indicates geometric treatment coverage but does not directly establish hemostatic efficacy. Future evaluations should therefore incorporate clinically relevant measures such as the Visual Inspection of Bleeding (VIBe) score \cite{lewis2017development}, together with automatic pressure-flow regulation and dynamic three-dimensional surface reconstruction.

\section{CONCLUSIONS}

We present a vision-guided, bimanual robotic framework for supervised autonomous hemorrhage-control experiments in low-grade splenic trauma. The system combines bleeding segmentation, markerless tool tracking, and tool--tissue contact classification to coordinate suction and monopolar cauterization with two dVRK-Si PSMs. Experiments on a novel \textit{ex vivo} hemorrhage model achieved average surface cauterization coverage of 82.9\% across seven distinct AAST Grade I--II injuries. This result demonstrates the feasibility of coordinated autonomous treatment. Future work will focus on three-dimensional surface reconstruction, improved contact detection under blood-pool and occlusion conditions, and closed-loop regulation based on direct measurements of bleeding reduction.

\printbibliography

@article{eastridge2012death,
  title={Death on the battlefield (2001--2011): implications for the future of combat casualty care},
  author={Eastridge, Brian J and Mabry, Robert L and Seguin, Peter and Cantrell, Joyce and Tops, Terrill and Uribe, Paul and Mallett, Olga and Zubko, Tamara and Oetjen-Gerdes, Lynne and Rasmussen, Todd E and others},
  journal={Journal of trauma and acute care surgery},
  volume={73},
  number={6},
  pages={S431--S437},
  year={2012},
  publisher={LWW}
}

@article{gallastegi2022intraoperative,
  title={Intraoperative deaths: who, why, and can we prevent them?},
  author={Gallastegi, Ander Dorken and Mikdad, Sarah and Kapoen, Carolijn and Breen, Kerry A and Naar, Leon and Gaitanidis, Apostolos and El Hechi, Majed and Pian-Smith, May and Cooper, Jeffrey B and Antonelli, Donna M and others},
  journal={Journal of Surgical Research},
  volume={274},
  pages={185--195},
  year={2022},
  publisher={Elsevier}
}

@article{isensee2021nnu,
  title={nnU-Net: a self-configuring method for deep learning-based biomedical image segmentation},
  author={Isensee, Fabian and Jaeger, Paul F and Kohl, Simon AA and Petersen, Jens and Maier-Hein, Klaus H},
  journal={Nature methods},
  volume={18},
  number={2},
  pages={203--211},
  year={2021},
  publisher={Nature Publishing Group US New York}
}

@inproceedings{isensee2024nnu,
  title={nnu-net revisited: A call for rigorous validation in 3d medical image segmentation},
  author={Isensee, Fabian and Wald, Tassilo and Ulrich, Constantin and Baumgartner, Michael and Roy, Saikat and Maier-Hein, Klaus and Jaeger, Paul F},
  booktitle={International conference on medical image computing and computer-assisted intervention},
  pages={488--498},
  year={2024},
  organization={Springer}
}

@inproceedings{xu2025dvrk,
  title={dvrk-si: The next generation da vinci research kit},
  author={Xu, Keshuai and Wu, Jie Ying and Deguet, Anton and Kazanzides, Peter},
  booktitle={2025 International Symposium on Medical Robotics (ISMR)},
  pages={185--191},
  year={2025},
  organization={IEEE}
}

@article{roshanov2021bleeding,
  title={Bleeding Independently associated with Mortality after noncardiac Surgery (BIMS): an international prospective cohort study establishing diagnostic criteria and prognostic importance},
  author={Roshanov, Pavel S and Eikelboom, John W and Sessler, Daniel I and Kearon, Clive and Guyatt, Gordon H and Crowther, Mark and Tandon, Vikas and Borges, Flavia Kessler and Lamy, Andre and Whitlock, Richard and others},
  journal={British journal of anaesthesia},
  volume={126},
  number={1},
  pages={163--171},
  year={2021},
  publisher={Elsevier}
}

@article{abhilash2020early,
  title={Early management of trauma: The golden hour},
  author={Abhilash, Kundavaram Paul Prabhakar and Sivanandan, A},
  journal={Current Medical Issues},
  volume={18},
  number={1},
  pages={36--39},
  year={2020},
  publisher={Medknow}
}

@article{xu2025bleedorigin,
  title={Bleedorigin: Dynamic bleeding source localization in endoscopic submucosal dissection via dual-stage detection and tracking},
  author={Xu, Mengya and Zhou, Rulin and Wang, An and Lyu, Chaoyang and Li, Zhen and Zhong, Ning and Ren, Hongliang},
  journal={arXiv preprint arXiv:2507.15094},
  year={2025}
}

@article{lewis2017development,
  title={Development and validation of an intraoperative bleeding severity scale for use in clinical studies of hemostatic agents},
  author={Lewis, Kevin M and Li, Qing and Jones, Drew S and Corrales, JoMichelle D and Du, Hongyan and Spiess, Philippe E and Menzo, Emanuele Lo and DeAnda Jr, Abe},
  journal={Surgery},
  volume={161},
  number={3},
  pages={771--781},
  year={2017},
  publisher={Elsevier}
}

@article{richter2021autonomous,
  title={Autonomous robotic suction to clear the surgical field for hemostasis using image-based blood flow detection},
  author={Richter, Florian and Shen, Shihao and Liu, Fei and Huang, Jingbin and Funk, Emily K and Orosco, Ryan K and Yip, Michael C},
  journal={IEEE Robotics and Automation Letters},
  volume={6},
  number={2},
  pages={1383--1390},
  year={2021},
  publisher={IEEE}
}

@inproceedings{huang2021model,
  title={Model-predictive control of blood suction for surgical hemostasis using differentiable fluid simulations},
  author={Huang, Jingbin and Liu, Fei and Richter, Florian and Yip, Michael C},
  booktitle={2021 IEEE International Conference on Robotics and Automation (ICRA)},
  pages={12380--12386},
  year={2021},
  organization={IEEE}
}

@inproceedings{pei2026synergistic,
  title={Synergistic bleeding region and point detection in laparoscopic surgical videos},
  author={Pei, Jialun and Zhou, Zhangjun and Guo, Diandian and Li, Zhixi and Qin, Jing and Du, Bo and Heng, Pheng-Ann},
  booktitle={Proceedings of the IEEE/CVF Conference on Computer Vision and Pattern Recognition},
  pages={1396--1405},
  year={2026}
}

@article{okamoto2019real,
  title={Real-time identification of blood regions for hemostasis support in laparoscopic surgery},
  author={Okamoto, Takayuki and Ohnishi, Takashi and Kawahira, Hiroshi and Dergachyava, Olga and Jannin, Pierre and Haneishi, Hideaki},
  journal={Signal, Image and Video Processing},
  volume={13},
  number={2},
  pages={405--412},
  year={2019},
  publisher={Springer}
}

@inproceedings{miao2024hemoset,
  title={Hemoset: The first blood segmentation dataset for automation of hemostasis management},
  author={Miao, Albert J and Lin, Shan and Lu, Jingpei and Richter, Florian and Ostrander, Benjamin and Funk, Emily K and Orosco, Ryan K and Yip, Michael C},
  booktitle={2024 International Symposium on Medical Robotics (ISMR)},
  pages={1--7},
  year={2024},
  organization={IEEE}
}

@article{ou2024autonomous,
  title={Autonomous blood suction for robot-assisted surgery: A sim-to-real reinforcement learning approach},
  author={Ou, Yafei and Soleymani, Abed and Li, Xingyu and Tavakoli, Mahdi},
  journal={IEEE Robotics and Automation Letters},
  volume={9},
  number={8},
  pages={7246--7253},
  year={2024},
  publisher={IEEE}
}

@article{zargarzadeh2025decision,
  title={From decision to action in surgical autonomy: Multi-modal large language models for robot-assisted blood suction},
  author={Zargarzadeh, Sadra and Mirzaei, Maryam and Ou, Yafei and Tavakoli, Mahdi},
  journal={IEEE Robotics and Automation Letters},
  volume={10},
  number={3},
  pages={2598--2605},
  year={2025},
  publisher={IEEE}
}

@article{ou2025learning,
  title={Learning autonomous surgical irrigation and suction with the da vinci research kit using reinforcement learning},
  author={Ou, Yafei and Tavakoli, Mahdi},
  journal={IEEE Transactions on Automation Science and Engineering},
  volume={22},
  pages={16753--16767},
  year={2025},
  publisher={IEEE}
}

@article{sogabe2023bleeding,
  title={Bleeding alert map (bam): The identification method of the bleeding source in real organs using datasets made on mimicking organs},
  author={Sogabe, Maina and Ishikawa, Kaoru and Takamatsu, Toshihiro and Takeuchi, Koh and Kanno, Takahiro and Fujimoto, Koji and Miyazaki, Tetsuro and Kawase, Toshihiro and Sato, Toshihiko and Kawashima, Kenji},
  journal={Array},
  volume={19},
  pages={100308},
  year={2023},
  publisher={Elsevier}
}

@article{hua2022automatic,
  title={Automatic bleeding detection in laparoscopic surgery based on a faster region-based convolutional neural network},
  author={Hua, Surong and Gao, Junyi and Wang, Zhihong and Yeerkenbieke, Palashate and Li, Jiayi and Wang, Jing and He, Guanglin and Jiang, Jigang and Lu, Yao and Yu, Qianlan and others},
  journal={Annals of translational medicine},
  volume={10},
  number={10},
  pages={546},
  year={2022}
}

@article{Mathisetal2018,
  title={DeepLabCut: markerless pose estimation of user-defined body parts with deep learning},
  author={Mathis, Alexander and Mamidanna, Pranav and Cury, Kevin M. and Abe, Taiga and Murthy, Venkatesh N. and Mathis, Mackenzie W. and Bethge, Matthias},
  journal={Nature Neuroscience},
  volume={21},
  number={9},
  pages={1281--1289},
  year={2018},
  publisher={Nature Publishing Group}
}

@article{NathMathisetal2019,
  title={Using DeepLabCut for 3D markerless pose estimation across species and behaviors},
  author={Nath, Tanmay and Mathis, Alexander and Chen, An Chi and Patel, Amir and Bethge, Matthias and Mathis, Mackenzie W.},
  journal={Nature Protocols},
  volume={14},
  number={7},
  pages={2152--2176},
  year={2019},
  publisher={Nature Publishing Group}
}

@inproceedings{tan2019efficientnet,
  title={EfficientNet: Rethinking model scaling for convolutional neural networks},
  author={Tan, Mingxing and Le, Quoc V.},
  booktitle={Proceedings of the 36th International Conference on Machine Learning},
  series={Proceedings of Machine Learning Research},
  volume={97},
  pages={6105--6114},
  year={2019},
  publisher={PMLR}
}

@misc{sekachev2020cvat,
  title={Computer Vision Annotation Tool (CVAT)},
  author={Sekachev, Boris and Manovich, Nikita and Zhiltsov, Maxim and Zhavoronkov, Andrey and Churin, Dmitry and Semenov, Yury and Chernova, Anna and Kustas, Andrey and Morozov, Alexey and Vasiliev, Andrey and Tereshchenko, Alexey and Kurilin, Kirill},
  year={2020},
  publisher={Zenodo},
  doi={10.5281/zenodo.4009388},
  url = {https://github.com/cvat-ai/cvat}
}


\end{document}